\documentclass[sigconf,natbib=false]{acmart}
\DeclareUnicodeCharacter{0301}{}
\usepackage{booktabs}
\usepackage{multirow} 
\usepackage{url}
\usepackage{xcolor}
\usepackage{bm}
\usepackage{algorithm}
\usepackage{algorithmic}

\definecolor{mygray}{gray}{.9}

\AtBeginDocument{%
  }

\setcopyright{acmcopyright}
\copyrightyear{2023}
\acmYear{2023}
\acmDOI{3581783.3611887}

\acmSubmissionID{1615}

\RequirePackage[
  datamodel=acmdatamodel,
  style=acmnumeric,
  ]{biblatex}

\begin{document}

\title{Synthesizing Long-Term Human Motions with Diffusion Models via Coherent Sampling}

\author{Zhao Yang}
\affiliation{%
  \institution{Gaoling School of Artificial Intelligence, Renmin University of China}
  \institution{Beijing Key Laboratory of Big Data Management and Analysis Methods}
  \streetaddress{59 Zhong Guan Cun Avenue}
  \city{Beijing}
  \country{China}
  \postcode{100872}
}
\email{yangyz1230@gmail.com}

\author{Bing Su}
\authornote{Corresponding author}
\affiliation{%
  \institution{Gaoling School of Artificial Intelligence, Renmin University of China}
  \institution{Beijing Key Laboratory of Big Data Management and Analysis Methods}
  \city{Beijing}
  \country{China}}
\email{subingats@gmail.com}

\author{Ji-Rong Wen}
\affiliation{%
  \institution{Gaoling School of Artificial Intelligence, Renmin University of China}
  \institution{Beijing Key Laboratory of Big Data Management and Analysis Methods}
  \city{Beijing}
  \country{China}}
\email{jrwen@ruc.edu.cn}

\renewcommand{\shortauthors}{Zhao Yang, Bing Su and Ji-Rong Wen}

\begin{abstract}
  Text-to-motion generation has gained increasing attention, but most existing methods are limited to generating short-term motions that correspond to a single sentence describing a single action. However, when a text stream describes a sequence of continuous motions, the generated motions corresponding to each sentence may not be coherently linked. Existing long-term motion generation methods face two main issues. Firstly, they cannot directly generate coherent motions and require additional operations such as interpolation to process the generated actions. Secondly,  they generate subsequent actions in an autoregressive manner without considering the influence of future actions on previous ones. To address these issues, we propose a novel approach that utilizes a past-conditioned diffusion model with two optional coherent sampling methods: Past Inpainting Sampling and Compositional Transition Sampling. Past Inpainting Sampling completes subsequent motions by treating previous motions as conditions, while Compositional Transition Sampling models the distribution of the transition as the composition of two adjacent motions guided by different text prompts. Our experimental results demonstrate that our proposed method is capable of generating compositional and coherent long-term 3D human motions controlled by a user-instructed long text stream. The code is available at \href{https://github.com/yangzhao1230/PCMDM}{https://github.com/yangzhao1230/PCMDM}.

\end{abstract}



\maketitle

\begin{figure}[ht]
\vspace{-0.2in}
\setlength{\belowcaptionskip}{0pt} 
\begin{center}
\includegraphics[width=0.8\linewidth]{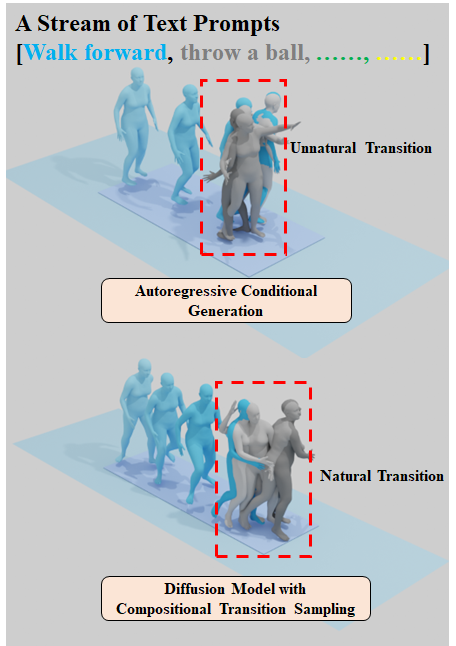}
\caption{Generating Coherent Long-Term 3D Human Motion from Text Streams. Existing autoregressive methods often lack natural transitions, leading to the use of alignment and interpolations between generated motions. In contrast, our proposed method employs diffusion models with coherent sampling methods, enabling the generation of smooth actions without the need for additional post-processing.}
\label{fig:Motivation}
\end{center}
\end{figure}

\vspace{-4mm}
\section{Introduction}
Synthesizing complex and realistic 3D human motions is of great significance in virtual reality, game production, film industry, human-machine interaction, etc. Motion capture techniques~\cite{romero2017embodied} have been widely applied to obtain motion data, but they are expensive and time-consuming because sophisticated equipment and professional actors are required to perform actions. It is desirable to develop motion generation models that can automatically generate demanded human motions as specified.

Early motion generation methods~\cite{petrovich2021action,  guo2020action2motion, cervantes2022implicit} can only generate motions from a fixed set of pre-defined action classes, which cannot meet highly flexible requirements in open real-world applications. A more convenient way to specify the target motion is by describing it in natural language. Text-to-motion generation~\cite{ahuja2019language2pose, guo2022generating, petrovich2022temos} aims to generate corresponding motions following any given language instructions. With the development of multi-modal pretraining ~\cite{sohn2015learning, huo2021wenlan} and generative models~\cite{sohl2015deep, ho2020denoising, kingma2013auto, goodfellow2014generative, kingma2018glow}, text-to-motion generation has made significant progress and attracted a lot of attention. However, most existing methods can only take a single sentence describing a single motion as the input prompt and generate an individual short-term motion.

In practice, a motion is usually not executed separately, but a series of motions are executed successively to perform an event or a complex activity. For example, as shown in Fig.~\ref{fig:Motivation}, the actor follows a stream of language instructions to complete one shot of a movie and the character in a game performs a series of actions to complete a task. When a stream of text prompts is fed into existing text-to-motion generation models, they can only be separately processed to generate a series of individual motions that are not coherent. Simple linear interpolations among motions cannot lead to natural and smooth transitions.

It is not trivial to extend existing text-to-motion generation models to tackle such long-term compositional motion generation for the following reasons. 1. Due to the semantic changes between different sentences, text streams differ significantly from the pre-training text prompts of these models. 2. It is not plausible to collect sufficient compositional motion-text data for pre-training these models since the number of motion compositions grows exponentially. 3. Non-autoregressive generative models can only generate motions with a fixed number of frames, but the number of sentences contained in streams is variable; autoregressive generative models rely on large amounts of compositional training data to model the transitions.

Relatively much fewer works have been devoted to long-term compositional motion generation. In~\cite{lee2022multiact}, although compositional motions can be generated, the input can only be multiple action labels rather than text prompts, and thus the motions for composition are limited to a pre-defined action set. In~\cite{athanasiou2022teach}, TEACH firstly tackles the problem of text-conditioned compositional motion generation, which employs the conditional variational autoencoder (VAE)~\cite{sohn2015learning} to successively generate motions by taking the last few frames of the previous motion as conditions. Although TEACH utilizes previous motion as a condition for generation, it still cannot directly generate coherent motions and requires alignment and interpolation between different motions.

Diffusion models have shown strong empirical performances and controllability in image and motion generation because the generation process is divided into multiple steps and each step only needs to fit a simple Gaussian distribution to reverse the forward diffusion. In this paper, we propose a novel approach for generating long-term compositional human motions from text streams using diffusion models. Specifically, we propose a past-conditioned diffusion model and two coherent sampling methods, namely Past Inpainting Sampling and Compositional Transition Sampling, that enable the direct generation of coherent motions without post-processing such as alignment and interpolation. 

The contributions of this paper are summarized as follows.
\begin{itemize}
\item To the best of our knowledge, we propose the first diffusion model-based method to tackle the problem of generating long-term compositional human motions from text streams. Our human motion diffusion model can be trained from general text-motion data, and no specific aligned text streams and long-term compositional human motions are required for training. To further improve the generation quality, we can also add a past-conditioned module to train on matched long-term text-action pairs.
\item We propose two sampling methods based on diffusion models that allow the model to directly generate coherent motions without the need for post-processing such as alignment and interpolation. These methods are Past Inpainting Sampling and Compositional Transition Sampling. Among them, Compositional Transition Sampling abandons the autoregressive generation paradigm, allowing adjacent motions to influence each other during generation and better aligning with the semantics of real long-term motions.
\item We conduct both quantitative and qualitative experimental evaluations to demonstrate the superiority of the proposed method over state-of-the-art text-instructed action composition methods.
\end{itemize}

\section{Related work}

We review relevant works in human motion generation and diffusion generative models.

{\bf Human motion generation.} Unconditional human motion generation~\cite{uncondition1, uncondition2} models human motion distributions without labels or guidance. Conditional human motion generation is gaining more interest. Some conditions are based on the motion itself, such as motion prediction~\cite{fragkiadaki2015recurrent, martinez2017human, hernandez2019human} or motion in-betweening~\cite{kaufmann2020convolutional, harvey2020robust} from prefix and suffix poses. Other conditions directly describe desired motion, such as action label conditioned~\cite{guo2020action2motion, cervantes2022implicit, petrovich2021action, lu2022action}, free-form text prompt conditioned~\cite{guo2022generating, petrovich2022temos, ahuja2019language2pose}, and music conditioned~\cite{li2022danceformer} motion generation. However, current methods struggle with generating coherent long-term motions composed of multiple actions. Recent work, such as \cite{lee2022multiact} and \cite{athanasiou2022teach}, introduce conditional VAE models to generate coherent compositional motions conditioned on multiple action labels or free-form language guidance. In contrast, we propose a diffusion model-based method that generates long-term compositional human motions from text streams, which does not require specific aligned text streams or long-term compositional human motions for training. We also introduce two sampling methods, Past Inpainting Sampling, and Compositional Transition Sampling, that enable direct generation of coherent motions without post-processing.

{\bf Diffusion generative models.} Diffusion models~\cite{sohl2015deep} add gradually increasing Gaussian noises to data and learn to reverse the perturbation via estimating the added noises. Recently, diffusion models have achieved success in various generative tasks, such as image generation~\cite{ho2020denoising, songscore}, audio generation~\cite{kongdiffwave}, text generation~\cite{lm, gong2022diffuseq}, story generation~\cite{story}, and molecule generation~\cite{geodiff, jo2022score, hoogeboom2022equivariant}. Large-scale pretrained diffusion models, such as DALL·E2~\cite{dalle}, Stable Diffusion~\cite{rombach2022high}, and Imagen~\cite{saharia2022photorealistic}, have dominated the field of text-to-image generation. Recently, some works applied diffusion models to 3D human motion generation, such as \cite{tevet2022human}, \cite{kim2022flame}, \cite{zhang2022motiondiffuse}, and \cite{yuan2022physdiff}. However, these methods cannot generate coherent long-term motions with multiple text prompts, which is our focus. \cite{liu2022compositional, du2023reduce} clarified the relationship between diffusion models and energy based models~\cite{du2019implicit, du2020compositional}, and propose methods of multi-conditioned compositional visual generation. We also consider the transition between adjacent actions as a composition of motions guided by different texts. 

\section{PRELIMINARY}
Inspired by non-equilibrium thermodynamics, diffusion models~\cite{sohl2015deep} define a Markov chain of the forward diffusion process $q$ to gradually add Gaussian noises to the real data distribution $\bm{x}_0 \sim q(\bm{x})$. The step sizes are controlled by a predefined variance schedule $\left\{\alpha_t \in(0,1)\right\}_{t=1}^T$. The forward process $q(\mathbf{x}_t \vert \mathbf{x}_{t-1})$ at each time step $t$ is:
\begin{equation}
\begin{aligned}
    q\left(x_t \mid x_{t-1}\right)=\mathcal{N}\left(x_t ; \sqrt{\alpha_t} x_{t-1},\left(1-\alpha_t\right) \mathbf{I}\right).
\end{aligned}
\end{equation}

After adding enough noise, the data distribution finally becomes equivalent to an isotropic Gaussian distribution.

DDPM~\cite{ho2020denoising} learns a prediction network denoted as $\boldsymbol{\epsilon}_{\theta}$ to reverse the forward diffusion process by predicting and removing the noise added in the corresponding forward step. Let $\beta_t = 1-\alpha_t$ and $\bar{\alpha}_t = \prod_{i=1}^t \alpha_i$. The denoising process $p(\mathbf{x}_{t-1} \vert \mathbf{x}_t)$ can also be reparameterized as a Gaussian distribution, which can be estimated by $\boldsymbol{\epsilon}_{\theta}$ and has a form of the following:
\begin{equation}
\begin{aligned}
p_\theta(\mathbf{x}_{t-1} \vert \mathbf{x}_t) &= \mathcal{N}(\mathbf{x}_{t-1}; \boldsymbol{\mu}_\theta(\mathbf{x}_t, t), \tilde{\beta}_t)  \\
\text{with} \quad \boldsymbol{\mu}_\theta(\mathbf{x}_t, t) &= \frac{1}{\sqrt{\alpha_t}} (\mathbf{x}_t - \frac{\beta_t}{\sqrt{1 - \bar{\alpha_t}}}\boldsymbol{\epsilon}_{\theta}(\mathbf{x}_t, t) ),
\end{aligned}
\label{eq:mu}
\end{equation}
where $\tilde{\beta}_t$ is a function of $\left\{\beta_t\right\}_{t=1}^T$. 

The learning objective of diffusion models is to approximate the mean $\boldsymbol{\mu}_\theta(\mathbf{x}_t, t)$ of the Gaussian distribution with respect to the reverse diffusion process. The mean for sampling $\mathbf{x}_{t-1}$ is estimated as the noisy data $\mathbf{x}_{t}$ minus the predicted noise $\boldsymbol{\epsilon}_{\theta}(\mathbf{x}_t, t)$ at step $t$. The variational lower bound (ELBO)~\cite{kingma2013auto} is adopted to minimize the negative log-likelihood of $p_\theta(\mathbf{x}_{0})$~\cite{ho2020denoising}, and the simplified objective can be written as a denoising objective:
\begin{equation}
\mathcal{L} = \mathbb{E}_{\mathbf{x}_0, \boldsymbol{\epsilon} \sim \mathcal{N}(0, 1), t} \Big[\|\boldsymbol{\epsilon} - \boldsymbol{\epsilon}_\theta(\mathbf{x}_t, t)\|^2 \Big].
\label{eq:objective}
\end{equation}

Once $\epsilon_\theta(x, t)$ is trained, we can use Eq.~(\ref{eq:mu}) to recover $\mu_\theta(x, t)$ and conduct ancestral sampling. Applying the properties of Markov chains, we can derive:
\begin{equation}
\label{eq:x0}
\begin{aligned}
    \boldsymbol{x}_0=\frac{x_t-\sqrt{1-\bar{\alpha}_t} \epsilon}{\sqrt{\bar{\alpha}_t}}   .
\end{aligned}
\end{equation}

We can also use Eq.~(\ref{eq:x0}) to recover clean data $\hat{\mathbf{x}}_\theta\left(\mathbf{x}_t, t\right)$.

For a conditional generation with diffusion models, we use classifier-free guidance~\cite{ho2022classifier}. Specifically, given condition $c$, we train a conditional model $\epsilon_\theta\left(\mathrm{x}t, c, t\right)$ and an unconditional model $\epsilon\theta\left(\mathrm{x}t, t\right)$ simultaneously. Then we can sample with:
\begin{equation}
\hat{\bm{\epsilon}}=s \cdot \bm{\epsilon}\theta\left(\mathrm{x}t, c, t\right)-(s-1) \cdot \bm{\epsilon}\theta\left(\mathrm{x}_t, t\right),
\label{eq:classifierfree}
\end{equation}
where $s$ is the guidance scale, and $c$ denotes the condition. 

\begin{figure*}[t]
\label{sec:overview}
\vspace{2pt}
\begin{center}
    \includegraphics[width=\linewidth]{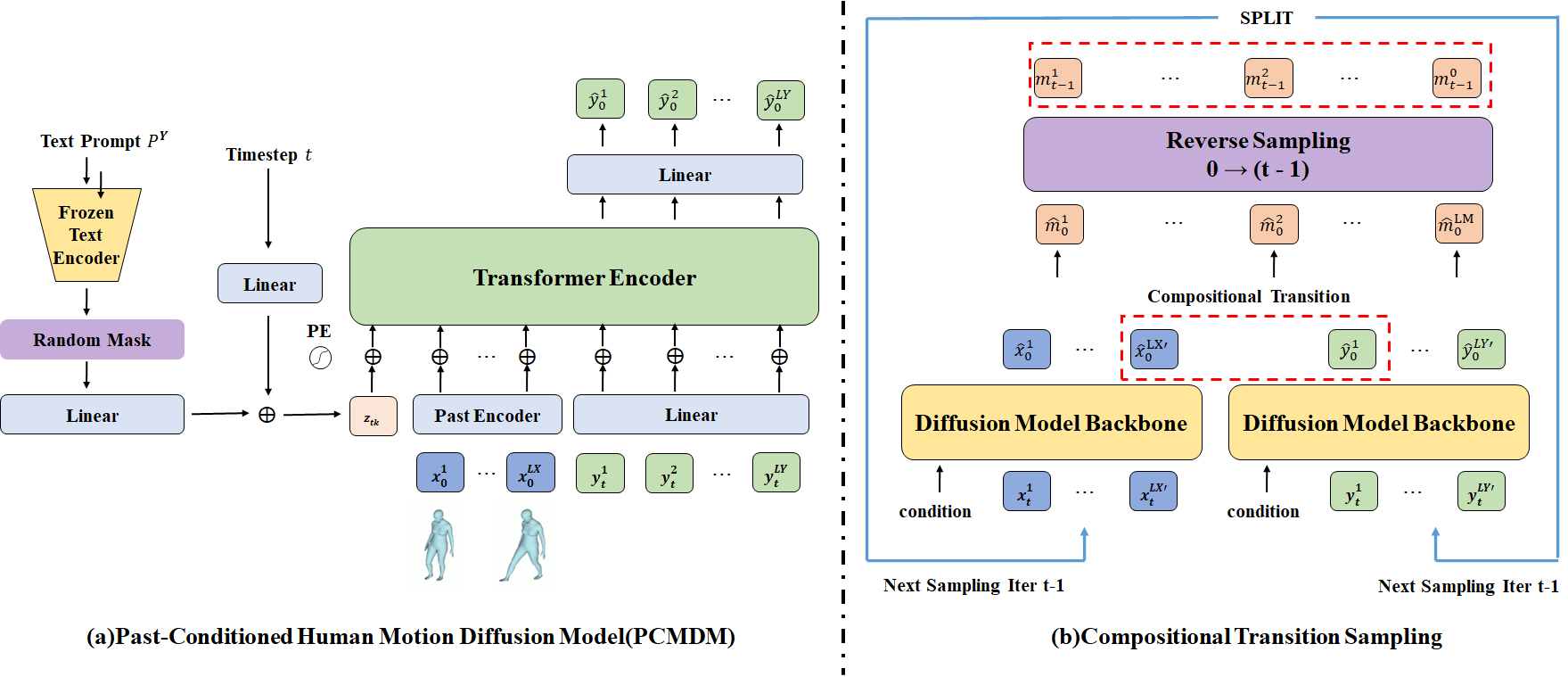}
\end{center}
\vspace{-6pt}
    \caption{ An overview of the proposed method. (Left): The architecture of our diffusion model, which is a Transformer Encoder used for predicting the denoised motion during the reverse process. The current text prompt is encoded using a frozen text encoder to obtain text features, which are added to the timestep embedding to obtain the token $z_{t}$. The last $h$ frames of the previous motion are encoded using the Past Encoder and are also used as token inputs. After being concatenated with the noisy second motion $y_t^{1: L_Y}$ at the $t$-th timestep, these tokens are fed into the Transformer Encoder to predict the denoised motion. (Right): Our proposed Compositional Transition Sampling. The compositional motion ${m}_t^{1: L_X + L_Y}$ is split into the two noisy motions ${x}_t^{1: L_X}$ and ${y}_t^{1: L_Y}$ at the $t$-th timestep. We add the first $L_{Tr / 2 }$ frames of ${y}_t^{1: L_Y}$ to ${x}_t^{1: L_X}$ to obtain ${x'}_t^{1: L_X'}$. We obtain ${y'}_t^{1: L_Y'}$ in a similar way. We then separately feed them into the diffusion model to predict the denoised motions $\hat{x'}_0^{1: L_X'}$ and $\hat{y'}_0^{1: L_Y'}$, respectively. They have an overlap of $L_{Tr}$ frames, which is the transition. By composing the overlapping part, the two motions are concatenated to form the denoised compositional motion $\hat{m}_0^{1: L_X+L_Y }$, which is further used to sample ${m}_{t-1}^{1: L_X + L_Y}$ at the $(t-1)$-th timestep.}
\label{fig:overview}
\vspace{-5pt}
\end{figure*}

\setlength{\abovedisplayskip}{0pt}
\setlength{\belowdisplayskip}{0pt}
\begin{figure}[ht]
    \vspace*{-\baselineskip} 
    \setlength{\abovecaptionskip}{0pt} 
    \setlength{\belowcaptionskip}{-10pt} 
    \begin{center}
        \includegraphics[width=1\linewidth]{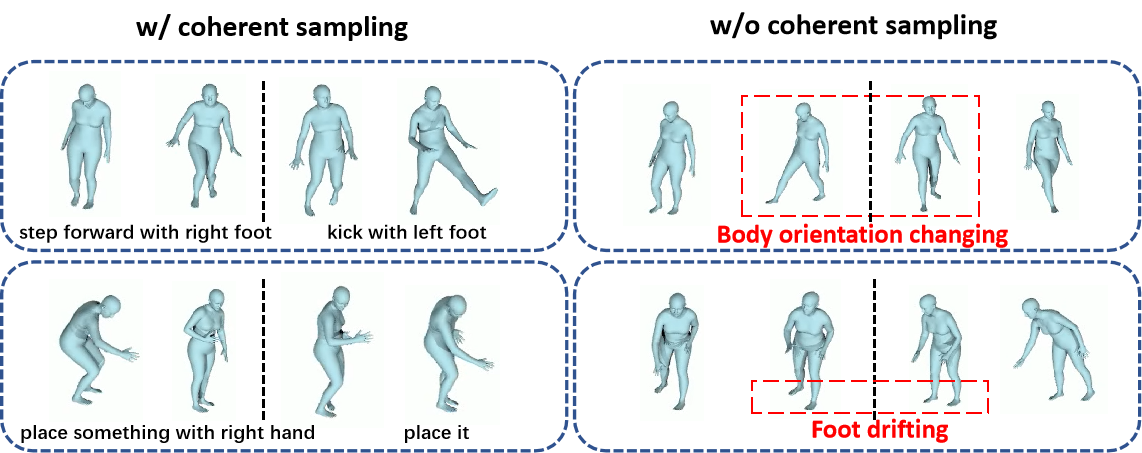}
        \caption{The conventional sampling methods cannot force adjacent motions to align. With our proposed Coherent Sampling methods, we can ensure that the generated motions are coherent and aligned.}
        \label{fig:discontin}
    \end{center}
\vspace{-0.1in}
\end{figure}

\section{METHOD}
\subsection{Problem Formulation}
We tackle the problem of text-conditioned long-term 3D human motion generation. Given an input stream of text prompts $\bm{S} = \{\bm{P}^1, \bm{P}^2, \cdots, \bm{P}^N\}$, where each prompt $\bm{P}^i$ describes a desired action and $N$ is the number of prompts, the goal is to generate a long-term realistic and coherent 3D human motion sequence $\bm{H} = \{\bm{X}^1, \bm{X}^2, \cdots, \bm{X}^N\}$ that sequentially contains the corresponding actions. Each sentence prompt $\bm{P}^i = \{\bm{w}_1^i, \bm{w}_2^i, \cdots, \bm{w}_{U_i}^i\}$ contains $U_i$ words, where  $\bm{w}_u^i$ is the $u$-th word, and the corresponding generated motion $\bm{X}^i = \{\bm{x}_1^i, \bm{x}_2^i, \cdots, \bm{x}_{L^i}^i\}$ has a predefined number $L^i$ of frames, where $\bm{x}_l^i$ is the $l$-th frame. Each frame $\bm{x}_l^i \in \mathbb{R}^{d}$, where $d$ is the dimensionality of the human motion representation. Following ~\cite{athanasiou2022teach}, motions are parametrized by SMPL body models~\cite{loper2015smpl} and are converted to a $6D$ rotation representation together with root translation. Therefore $d=135$ in our setting. We make the assumption in the following description of our method that there are only two segments of actions $\bm{H} = \{\bm{X}^1, \bm{X}^2\}$ (which can be extended to an arbitrary number of segments). To avoid confusion with the timestep $t$ in the diffusion model, we use $\bm{X} = \{\bm{x}^1, \bm{x}^2, \cdots, \bm{x}^{L_X}\}$ to represent $\bm{X}^1$ and $\bm{Y} = \{\bm{y}^1, \bm{y}^2, \cdots, \bm{y}^{L_Y}\}$ to represent $\bm{X}^2$. The corresponding stream of text Prompts is $\bm{S} = \{\bm{P}^{X}, \bm{P}^{Y}\}$.


\subsection{Overview}
Our proposed method consists of two main components: the first is the Past-Conditioned Human Motion Diffusion Model which can simultaneously incorporate text-guided conditions and the conditions of previous motions. The second consists of two coherent sampling methods that enable multiple motions to be seamlessly connected without the need for post-processing. The overview is shown in Fig.~\ref{fig:overview}, the left part (a) illustrates the architecture of the Past-Conditioned Human Motion Diffusion Model, while the right part (b) demonstrates one of our proposed sampling methods, the Compositional Transition Sampling strategy.


\subsection{Past-conditioned Human Motion Diffusion Model}
We model the forward diffusion process of the second action $Y$ as 
\begin{equation}
\begin{aligned}
    q\left(\bm{Y}_t\mid \bm{Y}_{t-1}\right)=\mathcal{N}\left(\sqrt{\alpha_t} \bm{Y}_{t-1},\left(1-\alpha_t\right) \bm{I}\right),
\end{aligned}
\end{equation}
where $t$ is the timestep of the diffusion model ($t=0$ means the clean motion data) and $\alpha_t$ are hyper-parameters to control the noise level of the forward process. We model the reverse distribution as a conditional distribution $p\left(\bm{Y}_{t-1} \mid \bm{Y}_{t}, {P}^Y, \bm{X}^{h}\right)$, where $\bm{X}^{h} = \bm{X}^{L_X - h + 1: L_X}$ and $h$ is a predefined hyperparameter that controls the last  $h$ frames of $\bm{P_X}$ as the conditional input.

Our model architecture, PCMDM, is inspired by MDM~\cite{tevet2022human} and TEACH~\cite{athanasiou2022teach}. We employ a Transformer Encoder as the backbone for text-conditioned motion generation while incorporating a Past Encoder (PC) to provide contextual information about the previous motion. Note that when the current motion is the first segment, we have $\bm{X}^{h}=\emptyset$. As shown in Fig.~\ref{fig:overview} (b), the text prompt $\bm{P}^Y$ is encoded through a pre-trained frozen text encoder. We obtain the token $\bm{z}_{t}$ by adding embeddings of text and timestep $t$. $\bm{X}^{h}$ is encoded through the Past Encoder. We concatenate these condition tokens and the current motion $\bm{Y}_t$ as the input to the Transformer Encoder. Note that our network ${Y}_{\theta}\left(\bm{Y}_t, t, \bm{P}^Y, \bm{X}^{h}\right)$ does not predict the noise $\boldsymbol{\epsilon}$ as in DDPM~\cite{ho2020denoising}, but directly predicts the clean human motion sequence $\bm{Y}_0$, which has proved to be effective in human motion generation in ~\cite{tevet2022human}. The training objective is:

\begin{equation}
\mathcal{L}=E_{\bm{Y}_0 \sim q\left(\bm{Y}_0 \mid \bm{P}^Y, {X}^{h} \right), t \sim[1, T]}\left[\left\|\bm{Y}_0-\boldsymbol{Y}_{\theta}\left(\bm{Y}_t, t, \bm{P}^Y, \bm{X}^{h}\right)\right\|_2^2\right].
\end{equation}

Once $\bm{Y}_{\theta}\left(\bm{Y}_t, t, \bm{P}^Y, \bm{X}^{h}\right)$ is trained, we can recover $\bm{\mu}_{\theta}\left(\bm{Y}_t, t, \bm{P}^Y, \bm{X}^{h}\right)$ as proven in Eq.~(\ref{eq:mu}) and Eq.~(\ref{eq:x0}). In the following text, for the sake of convenience, we may use either $\bm{Y}_{\theta}$ or $\bm{\mu}_{\theta}$ to represent the diffusion model backbone and we will ignore the conditions $\bm{P}^Y$ and $\bm{X}^{h}$.

\subsection{Sampling Strategies for Ensuring Motion Continuity}
With PCMDM, we can generate long-term motions in an autoregressive manner using a general diffusion model sampling method, similar to TEACH~\cite{athanasiou2022teach}. However, we have observed a problem of discontinuity between adjacent motions, as shown in Fig.~\ref{fig:discontin}. To address this issue, we have designed two improved sampling methods: Inpainting Sampling and Compositional Transition Sampling. These methods naturally align adjacent motions without the need for alignment and interpolation between motions.

\textbf{Inpainting Sampling.}
We cast the long-term 3D human motion generation as an image inpainting problem~\cite{saharia2022palette, lugmayr2022repaint, nichol2021glide}.  In inpainting, the objective is to predict missing pixels of an image using a know region as a condition, while in our case, we aim to predict future motion using the ending frames of past generated motion as a condition.

For a pair of adjacent text streams $\bm{S} = \{\bm{P}^{X}, \bm{P}^{Y}\}$, assuming that we have generated $\bm{X}_0$. The last $h$ frames of $\bm{X_0}$ are denoted by $\bm{x}_0^{{L}_X-h+1:L_X}$. We define $\bm{X^{ref}}=[\bm{x}_0^{{L}_X-h+1:L_X}; \bm{0}^{h+1: h+L_Y} ]$. We first sample ${\bm{Y}}_{T}$ of length $h+L_Y$ from the standard Gaussian distribution. We then generate a binary mask $\bm{m} = [\bm{1}^{1:h}; \bm{0}^{h+1: h+L_Y}]$ where positions with respect to history frames $\bm{x}_0^{{L}_X-h+1:L_X}$ are set to 1 and other positions are set to 0.

The difference with the original sampling step is that we overwrite the masked history motion (where $\bm{m}=1$) with the reference motion $\bm{x}_0^{{L}_X-h+1:L_X}$ at each step before sampling. At the $t$-th step, the reverse sampling process to generate ${\bm{Y}}_{t-1}$ given ${\bm{Y}}_{t}$ is as follows:     
\vspace{-0.1in}
\begin{subequations}
\begin{align}
    {\hat{\bm{Y}}}_{0} &= \bm{Y}_{\theta}\left(\bm{Y}_t, t\right), \\
   \hat{\bm{Y}}_{0} &= \bm{m} \odot \bm{X}^{ref} + (1-\bm{m}) \odot {\hat{\bm{Y}}}_{0}.
\end{align} \label{eq:ourStep}
\end{subequations}

To sample ${\bm{Y}}_{t-1}$, we can use the modified predicted ${\hat{\bm{Y}}}_{0}$. This means that ${{\bm{Y}}}_{t-1}$ is sampled while conditioning on the known human motion generated for the previous sentence. This approach called Past Inpainting sampling, connects the generated motion ${\bm{Y}}_0$ with the previous motion ${\bm{X_0}}$. 

\textbf{Compositional Transition Sampling.} TEACH and Past Inpainting Sampling assume that only previously generated actions affect subsequent ones. However, in real continuous motion, previous and future actions may have mutual influences. For example, the transitions from "running, jumping" to "running, sitting down" have significant semantic differences, indicating that autoregressive generation is not sufficient. Therefore, we propose a one-shot sampling method that models the distribution of transitions as the product of two adjacent different conditional action distributions. This method allows for the simultaneous consideration of the mutual interaction between previous and future actions in generating long-term human motion.

The idea of sampling from distributions that are the product of different experts originally comes from energy-based models(EBM)~\cite{du2019implicit, du2020compositional}. ~\cite{liu2022compositional} builds on this idea by exploring the connection between diffusion models and energy-based models, and proposes a method of directly adding the weighted noise predicted by diffusion models trained under different conditions to generate images with multiple instructions through the product of these distributions:
\begin{equation}
\label{eq:img_comp}
\begin{aligned}
    \hat{\mu}\left(\boldsymbol{x}_t, t\right)=\sum_{i=1}^n w_i\left(\mu_\theta\left(\boldsymbol{x}_t, t \mid \boldsymbol{c}_i\right)\right),
\end{aligned}
\end{equation}
where $\boldsymbol{c}_i$ represents different conditions, and $\boldsymbol{w}_i$ represents the weight of each condition's impact. In theory, the distribution sampled in this way is only equivalent to the product of different distributions when $t=0$ and $t=T$, theoretically. However, the experimental results in generating composite conditional images have shown good performance.

{\centering
\begin{figure}[t]
\begin{minipage}{\linewidth}
  \begin{algorithm}[H]
    \caption{Compositional Transition Sampling}
    \label{alg:compositional}
    \begin{algorithmic}[1]
    \STATE \textbf{Require} Human Motion Diffusion Model: $\mu_\theta\left(\bm{X}_t, t \mid P^X\right)$ and $\mu_\theta\left(\bm{Y}_t, t \mid P^Y\right)$; length of $\bm{X}$: $L_X$; length of $\bm{Y}$: $L_Y$; number of transition frames $\bm{L}_{Tr}$;  covariances $\tilde{\beta}_t$; weight function $w$; Operator \textbf{SPLIT}\\
    \STATE Initialize ${\bm{M}}_{T} \sim \mathcal{N}(\bm{0}, \bm{I})$ \\
    \FOR{$t = T, \ldots, 1$}
        \STATE $\bm{X}'_t, \bm{Y}'_t \gets \textbf{SPLIT}({\bm{M}}_{t})$
        \STATE $\mu_M^i \gets w(i)\mu_\theta^i(\bm{X}'_t, t \mid P^X) + (1-w(i))\mu_\theta(\bm{Y}'_t, t \mid P^Y)$
        \STATE ${\bm{M}}_{t-1} \sim \mathcal{N}(\bm{\mu}_{M}, \tilde{\beta}_t)$


    \ENDFOR
    \STATE Return ${\bm{M}}_{0}$
    \end{algorithmic}
  \end{algorithm}
\end{minipage}
\vspace{-.2cm}
\end{figure}
}

Our goal is to generate multiple actions that can be smoothly connected. It is non-trivial to directly apply this method to long-term human motion generation, because image generation applies all conditions to the same image canvas, while motion only exhibits such combined condition effects in relatively short transitions. Therefore, we define the transition between two adjacent actions $X$ and $Y$:

\begin{equation}
\begin{aligned}
    \bm{Tr} = \{\bm{x}^{L_X-{L_T/2} + 1}, \cdots, \bm{x}^{L_X}, \bm{y}^{1}, \cdots, \bm{y}^{L_T/2}\},
\end{aligned} 
\end{equation}
 where $L_X, L_Y$ and $L_T$ are the lengths of $\bm{X},\bm{Y}$and $\bm{Tr}$, respectively. To facilitate modeling, we add the length of $L_T/2$ to the end of $\bm{X}$ and the beginning of $\bm{Y}$, resulting in $\bm{X}'$ and $\bm{Y}'$:
\begin{equation}
\begin{aligned}
  \bm{X'} & = \{\bm{x}^{1}, \cdots, \bm{x}^{L_X}, \bm{y}^{1}, \cdots, \bm{y}^{L_T/2}\},\\
  \bm{Y'} & = \{\bm{x}^{L_X-L_T/2+1}, \cdots, \bm{x}^{L_X}, \bm{y}^{1}, \cdots, \bm{y}^{L_Y}\}.
\end{aligned} 
\end{equation}

We represent the whole motion obtained by concatenating $\bm{X}$ and $\bm{Y}$ as $\bm{M} = [\bm{X}; \bm{Y}]$. We define an operator \textbf{SPLIT} to directly obtain $\bm{X}'$ and $\bm{Y}'$ from $\bm{M}$:
\begin{equation}
\begin{aligned}
    \textbf{SPLIT}(\bm{M}) = (\bm{X'}, \bm{Y'}).
\end{aligned} 
\end{equation}

Note that we have $\bm{X'} \cap \bm{Y'} = \bm{Tr}$. The Transition between adjacent actions should contain semantic information that simultaneously captures both of their characteristics. Thus we assume that the distribution of ${Tr}$ is the product of the distributions of $q^{X^{\prime}}(\bm{x})$ and $q^{Y^{\prime}}(\bm{x})$. Therefore we have: 
\begin{equation}
\begin{aligned}
    q^{Tr}(\bm{x})=\frac{1}{\bm{Z}} q^{X^{\prime}}(\bm{x}) q^{Y^{\prime}}(\bm{x}), \quad Z=\int q^{X^{\prime}}(\bm{x}) d \bm{x} q^{Y^{\prime}}(\bm{x})d \bm{x},
\end{aligned} 
\end{equation}
where $Z$ is a normalization constant used to ensure that the integral of the probability distribution equals 1. Intuitively, the meaning of this distribution is that the high probability region of $q^{Tr}(\bm{x})$ is also a high probability region under the probability distributions of both $q^{X^{\prime}}(\bm{x})$ and $q^{Y^{\prime}}(\bm{x})$. In this way, we have defined the product of distributions at the transition, and therefore, we can use the method described in Eq.~(\ref{eq:img_comp}) to sample the transition:
\begin{equation}
\begin{aligned}
    \hat{\mu}_{Tr}\left(\bm{Tr}_t, t\right)=(\mu_\theta(\bm{X}'_t,t \mid P^X) + \mu_\theta(\bm{Y}'_t, t \mid P^Y))/2.
\end{aligned} 
\end{equation}

We assume that the influence of the preceding and subsequent actions on the transition is equally important. Therefore, we set the weight coefficients $w$ to $\frac{1}{2}$.

In the previous discussion, we consider the transition positions of two actions, and we can obtain the composed transition by simply adding the weighted sum of $\mu_\theta$ for each motion. For other non-overlapping areas, we can directly use the backbone PCMDM to predict $\mu_\theta$ for single-text-guided sampling. The sampling formula for $\bm{M}$ as a whole can be written as:
\begin{equation}
\begin{aligned}
    \mu_M^i = w(i)\mu_\theta^i(\bm{X}'_t, t \mid P^X) + (1-w(i))\mu_\theta(\bm{Y}'_t, t \mid P^Y),
\end{aligned} 
\end{equation}
where $\mu_M^i$ represents the sampling mean at the $i$-th frame, and $w(i)$ is a time-dependent weight function used to determine the position of the $i$-th frame:
\begin{equation}
\begin{aligned}
    w(i)=\begin{cases}
    1 & \text{if } i \leq L_X - {L_T}/2 \text{ or } i > L_Y + {L_T}/2 \\
    1/2 & \text{otherwise}
    .
\end{cases}
\end{aligned} 
\end{equation}

The complete sampling process is in Alg.~\ref{alg:compositional}. With Compositional Transition Sampling, we can directly obtain coherent actions without the need for alignment and interpolation.

\section{Experiments}
\begin{table*}[t]
\centering
\begin{tabular}{lcccc}
\toprule
Method & FID\,$\downarrow$ & R-Precision(TOP 3)\,$\uparrow$ & MultimodalDist\,$\downarrow$ & diversity\,$\rightarrow$ \\ 
\midrule
Real & $0.009^{\pm0.001}$ &$0.773^{\pm0.002}$ & $21.860^{\pm0.006}$ & $15.034^{\pm0.078}$ \\
\midrule
TEACH-Independent~\cite{athanasiou2022teach} & $12.256^{\pm.0.284}$ & $\underline{0.816}^{\pm.0.004}$ & ${21.874}^{\pm.0.115}$ & ${13.905}^{\pm.0.066}$ \\
TEACH-Joint~\cite{athanasiou2022teach} & $13.084^{\pm.0.284}$ & ${0.783}^{\pm.0.008}$ & ${22.218}^{\pm.0.103}$ & ${13.624}^{\pm.0.094}$ \\
TEACH~\cite{athanasiou2022teach} & $7.312^{\pm.0.0190}$ &$\mathbf{0.864}^{\pm.0.005}$ & $\mathbf{21.017}^{\pm.0.078}$ & ${14.2483}^{\pm.0.070}$ \\
\midrule
MDM~\cite{tevet2022human} & ${7.476}^{\pm0.232}$ &$0.770^{\pm0.0.008}$ & $22.020^{\pm0.098}$ & $13.876^{\pm.0.071}$ \\
MDM w/ Inpainting Sampling & $7.411^{\pm.0.242}$ &${0.772}^{\pm.0.005}$ & ${22.038}^{\pm.0.109}$ & ${13.788}^{\pm.0.083}$ \\
MDM w/ Compositional Sampling & $7.057^{\pm.0.232}$ &${0.782}^{\pm.0.006}$ & ${21.868}^{\pm.0.110}$ & ${14.112}^{\pm.0.0871}$ \\
\midrule
PCMDM & $\underline{5.396}^{\pm.0.187}$ &${0.775}^{\pm.0.010}$ & ${21.653}^{\pm.0.120}$ & $\underline{14.5338}^{\pm.0.063}$ \\
PCMDM w/ Inpainting Sampling & $5.431^{\pm.0.176}$ &${0.778}^{\pm.0.008}$ & ${21.646}^{\pm.0.126}$ & ${14.501}^{\pm.0.084}$ \\
PCMDM w/ Compostional Sampling & $\mathbf{5.242}^{\pm.0.131}$ &${0.799}^{\pm.0.007}$ & $\underline{21.412}^{\pm.0.087}$ & $\mathbf{14.652}^{\pm.0.068}$ \\
\bottomrule

\end{tabular}
\caption{
\textbf{Comparison with Previous State of the Art.} We report the performance of TEACH, MDM, and our PCMDM with our proposed sampling methods. \textit{w/ Inpainting Sampling''} and \textit{w/ Compositional Sampling''} refer to using our proposed novel sampling methods on diffusion models. We run all the evaluations 10 times. \textbf{Bold} indicates the best result, \underline{underline} indicates the second-best result, $\pm$ indicates a 95 confidence interval, and $\rightarrow$ indicates that closer to real is better.
}
\label{table:compare}
\vspace{-0.2in}
\end{table*}

\begin{table}[t]
\captionsetup[table]{skip=-5pt} 
\centering
\begin{tabular}{lcc}
\toprule
\multirow{2}{*}{\textbf{Methods}} & \multicolumn{2}{c}{Transition Dist$\downarrow$}\\
& w/ align. & w/out align.\\
\midrule
TEACH-Independent~\cite{athanasiou2022teach} & 0.151 & 0.177 \\
TEACH-Joint~\cite{athanasiou2022teach} & 0.107 & 0.122 \\
TEACH~\cite{athanasiou2022teach} & 0.107 & 0.122 \\   
\midrule
MDM~\cite{tevet2022human} & n/a   & 0.248 \\    
MDM w/ Inpainting Sampling & n/a   & 0.116 \\  
MDM w/ Compostional Sampling & n/a   & $\mathbf{0.009}$ \\  
\midrule
PCMDM & n/a   & 0.119 \\  
PCMDM w/ Inpainting Sampling & n/a   & 0.090 \\  
PCMDM w/ Compostional Sampling & n/a   & $\underline{0.014}$ \\  
\bottomrule
\end{tabular}
\caption{We measure the transition distance for generated samples. "w/ align." indicates that the second motion is translated and rotated to match the first motion. Our method achieves significantly better results than other methods without the need for alignment.}
\label{tab:transition distance}
\vspace{-0.4in}
\end{table}

\subsection{Dataset}


In traditional text-to-motion datasets, each motion sequence is associated with a single text. However, the BABEL dataset~\cite{punnakkal2021babel} provides more fine-grained annotations for long-term motion sequences, with each subsequence corresponding to a specific text annotation. In total, there are 10881 motion sequences, with 65926 subsequences and the corresponding textual labels. Following TEACH~\cite{athanasiou2022teach}, we create subsequence pairs by taking adjacent pairs of subsequences from the long sequences. For example, if the original data contains a motion sequence ['walk', 'sit down', 'wave right hand'], we can construct two subsequence pairs from it: ['walk', 'sit down'] and ['sit down', 'wave right hand']. The BABEL dataset itself comes with pre-defined train and test splits, and we adopt the default data partitioning of BABEL. After dividing the long sequences into subsequence pairs, there are approximately 15.7k and 5.7k pairs in the training and testing sets respectively.

Then we will elaborate on the specific training details for conducting training on such a dataset. Assuming we have a training data example with a text stream of ['walk', 'sit down'], and the corresponding motion sequence consist of two subsequences aligned with the prompts. Since MDM~\cite{tevet2022human} does not have a Past Encoder, its training objective is to independently generate the corresponding subsequences for each prompt. On the other hand, TEACH~\cite{athanasiou2022teach} and PCMDM have a Past Encoder, so the training objective for the "walk" subsequence is to generate the motion based on the prompt alone, while for the "sit down" subsequence, the training objective is to generate the second motion by taking both the previous motion subsequence and the prompt "sit down" as conditions to feed into the model.


\subsection{Evaluation Metrics}
TEACH uses Average Positional Error (APE) and Average Variational Error (AVE) for evaluation, both of which calculate the distance between generated motions and ground truth motions. Such metrics are not suitable for evaluating generative models. Following recent work~\cite{guo2022generating, tevet2022human} on text-guided human motion generation, we use other alternative metrics to evaluate our method.

\begin{figure}[ht]
\vspace{-0.1in}
\begin{center}
\includegraphics[width=1\linewidth]{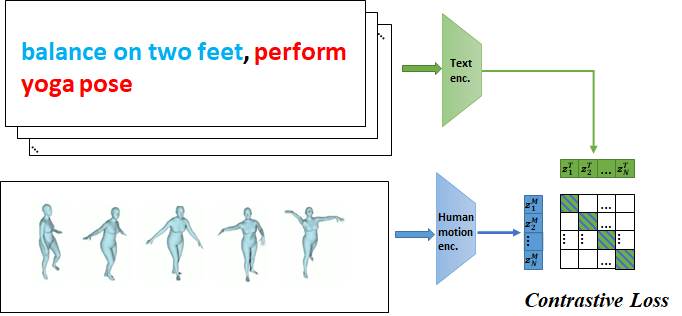}
\setlength{\abovecaptionskip}{0pt} 
\setlength{\belowcaptionskip}{-10pt} 
\caption{\textbf{The training architecture of MotionClip.}}
\label{fig:motionclip}
\end{center}
\vspace{-0.1in}
\end{figure}

To this end, following~\cite{kim2022flame}, we adapt CLIP~\cite{clip} and train a MotionCLIP with a motion encoder and a text encoder on the training set through contrastive learning. We concatenate adjacent motion pairs as long-term motion and separate adjacent text descriptions with a comma as long-term text. These are then used as inputs to the MotionCLIP, as shown in Fig.~\ref{fig:motionclip}. These two encoders can be used to evaluate the following metrics:
\textbf{Frechet Inception Distance (FID)} represents the distribution divergence from generated samples to real data. A lower value implies better FID results.
\textbf{R Precision} is calculated by putting the ground-truth text and a set of randomly selected mismatched descriptions from the test set into a pool for each generated motion. We calculate the Euclidean distance between the motion feature and the text feature of each description in the pool and count the average top-3 accuracy.
\textbf{Diversity} measures the variance of the generated motions across all generated human motions.
\textbf{Matching Scores} calculate the distance between the motion feature and the corresponding text feature extracted by the encoders.

In addition, a key metric for evaluating the quality of generated long-term human motion is the coherence between adjacent actions. Therefore, we use the Transition Distance proposed by~\cite{athanasiou2022teach}.
\textbf{Transition Distance} quantifies the degree of discontinuity by calculating the average transition distance, which is defined as the Euclidean distance between the body poses of the last frame of the preceding action and the first frame of the subsequent action. Our method does not require alignment and outperforms previous methods even without it.

\subsection{Implementation details}
We trained our human motion diffusion models (both MDM and PCMDM) for 50,000 steps using the AdamW optimizer~\cite{kingma2014adam, loshchilov2017decoupled} with a fixed learning rate of $10^{-4}$. The minibatch size was set to 32 for both training and evaluation. The MDM architecture is the same as in~\cite{tevet2022human}, and our PCMDM is an MDM with a past-conditioned encoder, which is a linear layer. During testing, we randomly selected 1000 text prompts and their corresponding lengths from the test set, generated continuous motion pairs, and calculated the metrics for these 1000 generated data points for the first four metrics. We repeated this experiment 10 times. For the Transition Distance, we conducted a single experiment on the entire test set following~\cite{athanasiou2022teach}. We conducted experiments on various types of GPUs, mainly NVIDIA Tesla V100-32GB GPUs and NVIDIA Tesla A800-80GB GPUs. We used the code and model weights provided for the Teach baselines experiments.


\begin{table*}[t]
\centering
\begin{tabular}{lcccc}
\toprule
Method & FID\,$\downarrow$ & R-Precision(Top 3)\,$\uparrow$ & MultimodalDist\,$\downarrow$ & diversity\,$\rightarrow$ \\
\midrule
REAL & $0.009^{\pm0.001}$ &$0.773^{\pm0.002}$ & $21.860^{\pm0.006}$ & $15.034^{\pm0.078}$ \\
\midrule
inpainting frames $h = 2$ & $5.431^{\pm0.176}$ &$0.778^{\pm0.008}$ & $\mathbf{21.646}^{\pm0.126}$ & $\mathbf{14.501}^{\pm.0.084}$ \\
inpainting frames $h = 4$ & $\mathbf{5.416}^{\pm0.179}$ &$0.778^{\pm0.009}$ & $21.675^{\pm0.115}$ & $14.483^{\pm0.091}$ \\
inpainting frames $h = 8$ & ${5.491}^{\pm0.164}$ &$\mathbf{{0.779}}^{\pm0.006}$ & ${21.692}^{\pm0.093}$ & $14.470^{\pm0.085}$ \\
inpainting frames $h = 12$ & {$5.613^{\pm.0.171}$} &$0.775^{\pm0.007}$ & $21.740^{\pm0.089}$ & ${14.436}^{\pm0.121}$ \\
\midrule
transition frames $L_{Tr} = 2$ & $\mathbf{5.242}^{\pm.0.131}$ &${0.799}^{\pm.0.007}$ & ${21.412}^{\pm.0.087}$ & $\mathbf{14.652}^{\pm.0.068}$ \\
transition frames $L_{Tr} = 4$ & ${5.269}^{\pm0.122}$ &$\mathbf{0.804}^{\pm0.0.006}$ & $\mathbf{21.399}^{\pm0.084}$ & $14.636^{\pm.0.082}$ \\
transition frames $L_{Tr} = 8$ & ${5.285}^{\pm0.151}$ &$0.803^{\pm0.0.007}$ & $21.457^{\pm0.089}$ & $14.639^{\pm.0.102}$ \\
transition frames $L_{Tr} = 12$ & ${5.627}^{\pm0.174}$ &$0.787^{\pm0.0.007}$ & $22.628^{\pm0.092}$ & $14.600^{\pm.0.067}$ \\
\bottomrule
\end{tabular}

\caption{
    \textbf{Ablation study on the number of inpainting frames in inpainting sampling and transition frames in compositional transtion sampling.}
    We ablate the performance with different $h$.
}
\label{tab:abla frames}
\vspace{-0.3in}
\end{table*}


\subsection{Comparison with State-of-the-art}
\label{sec:sota}
TEACH is the only existing work that focuses on long-term 3D human motions from text streams, so we mainly compared our proposed method with all the baseline models proposed in TEACH. Among them, TEACH-Independent is a VAE model trained directly on single text-action pairs, TEACH-Joint is a VAE trained on longer sequences by connecting adjacent text-action pairs, and TEACH incorporates information from previous actions into the training of subsequent actions using a Past-condition module similar to our proposed method. MDM is a diffusion model trained directly on single text-action pairs, and PCMDM is a diffusion model with previous action information injected.

As shown in Tab.~\ref{table:compare}, our proposed PCMDM with Compositional Transition Generation achieves state-of-the-art results on FID and diversity, indicating that our method can generate high-quality long-term human motion. Our method performs slightly worse than Teach on R-precision and MultimodalDist, which may be due to the fact that we used a smaller text encoder, Clip base, compared to Teach's DistilBert. It is also possible that our MotionCLIP was not fed with sufficient training data, resulting in encoder features that are not good enough, as our metrics are closer to real-world test data.

In addition, as shown in Tab.~\ref{tab:transition distance}, we can see that our method's generated actions are much more consistent than the baselines. It is worth noting that our method outperformed the baseline by a factor of 10 on Transition Distance without any post-processing. This indicates that our proposed method can generate more coherent and smooth long-term human motion sequences, which is crucial for real-world applications such as virtual reality and robotics.

\vspace{-0.2in}
\begin{figure}[ht]
\label{fig:abla_scale}
\begin{center}
\includegraphics[width=1\linewidth]{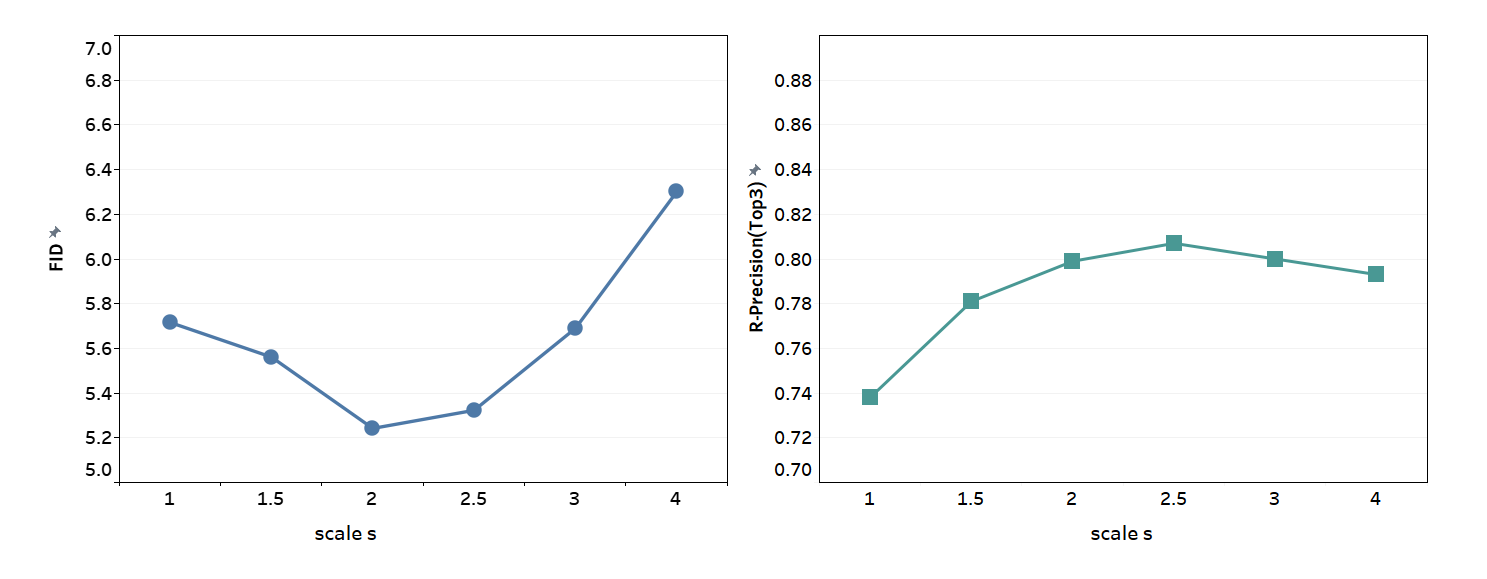}
\caption{\textbf{Abalation study on text condition scale $s$.} We observe $s = 2$ obtains the best performance.}
\label{fig:condition_scale}
\end{center}
\vspace{-0.2in}
\end{figure}

\begin{figure*}[t]
\vspace{-0.2in}
\begin{center}
\includegraphics[width=0.9\linewidth]{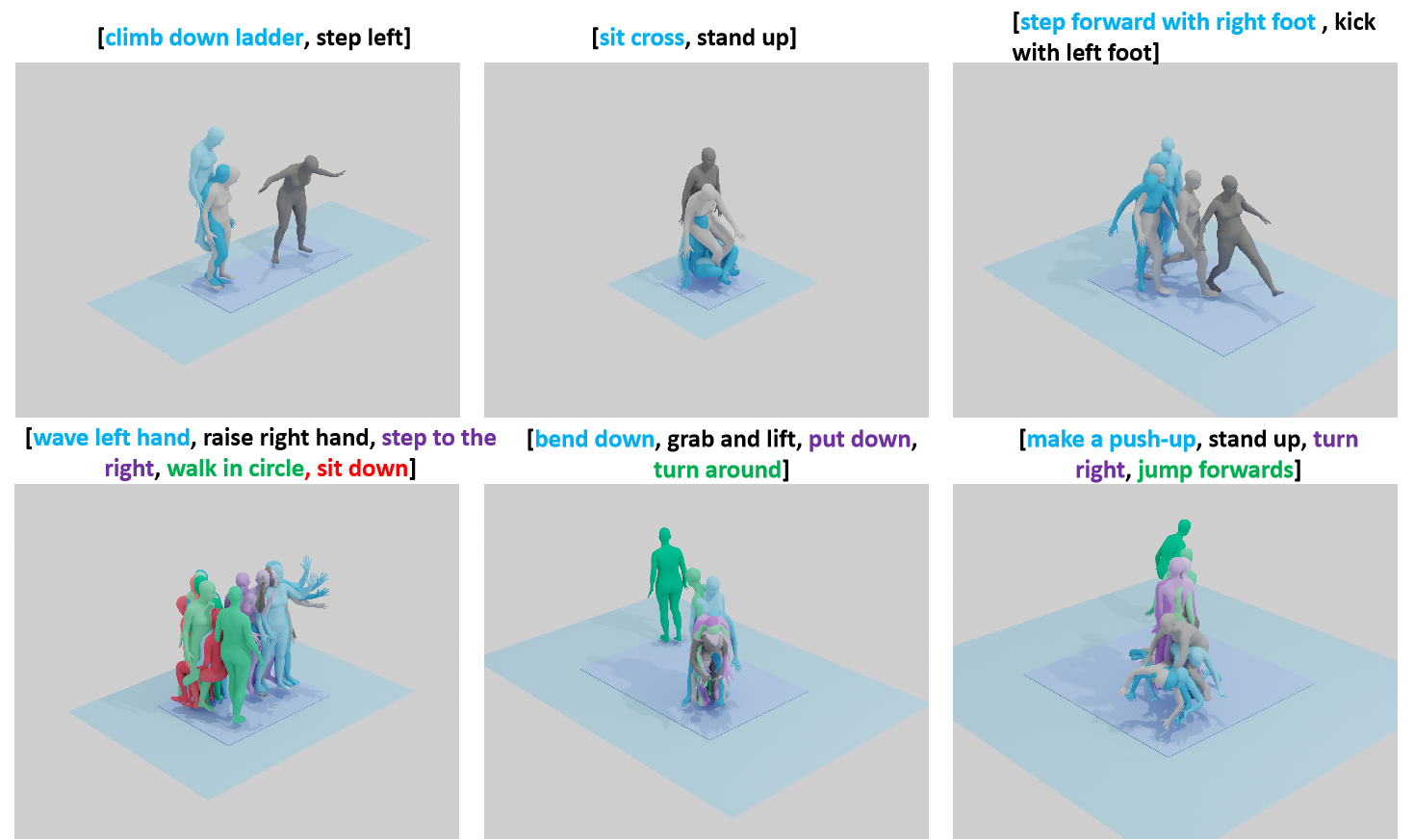}
\vspace{-0.1in}
\caption{\textbf{Qualitative results of our proposed method on long-term human motion generation.} Our method generates coherent and diverse 3D human motions for streams with more than two prompts. Different colored motions represent different text prompts. On the left, we show the overall motion in a single image to observe positional changes. On the right, we display the motion frame by frame, making it easier to see details of the body pose changes.}
\label{fig:qualitative}
\end{center}
\vspace{-0.1in}
\end{figure*}

\subsection{Ablation Study}
We provide an extensive analysis of the proposed PCMDM with a special sampling method, focusing on the influence of adjustable but important parameters: (1) inpainting frames and transition frames and (2) text condition scale.

{\bf Effect of inpainting frames $h$ and transition frames $L_{Tr}$.} As shown in Fig.~\ref{tab:abla frames}, when the values are set to smaller values, the performance is relatively better. This may be because using too many frames to guide coherent generation can harm the semantic information of individual motions.

{\bf Effect of text condition weight $s$.} The conditional weight coefficient $s$ in Eq.(\ref{eq:classifierfree}) is a crucial parameter in the classifier-free guidance, where a larger value leads to higher fidelity to the condition but may decrease the faithfulness to the true distribution of the original data. Through experiments, we found that $s=2$ works best for our task.

\subsection{Qualitative Results of Long-Term Motion Generation}

In Section~\ref{sec:sota}, we simplify the challenging task of generating coherent long-term 3D human motion by following the setting of~\cite{athanasiou2022teach}, where the text stream contains two text prompts. However, our method is easily extendable to generate longer coherent 3D human motions for streams with more than two text prompts. In Fig.~\ref{fig:qualitative}, we present several examples of our method generating motions with two or more challenging prompts. Our results demonstrate that our approach is capable of generating highly realistic 3D human motions, with smooth and natural transitions between successive actions.
Fig.~\ref{fig:qualitative} presents our generated motions in two different formats. On the left, we render the motion onto a single image, making it easier to observe motions with positional changes in space. On the right, we display the long-term human motion frame by frame, making it easier to see details of the body pose changes.

\section{Conclusion}
In this paper, we propose a novel approach for generating long-term 3D human motions from text streams, by utilizing a past-conditioned human motion diffusion model and two coherent sampling methods called Past Inpainting and Transition Composition Sampling. Our approach generates coherent long-term motions corresponding to each sentence in the input stream without the need for post-processing. Our experimental results demonstrate that our proposed method is able to generate high-quality 3D human motions controlled by a user-instructed long text stream. 

\section*{Acknowledgements}
This work was supported in part by the National Natural Science Foundation of China No. 61976206 and No. 61832017, Beijing Outstanding Young Scientist Program NO. BJJWZYJH012019100020098, Beijing Academy of Artificial Intelligence (BAAI), the Fundamental Research Funds for the Central Universities, the Research Funds of Renmin University of China 21XNLG05, and Public Computing Cloud, Renmin University of China.

\printbibliography

\balance
 
\newpage
\appendix

\section{APPENDIX}

We report more experimental results and more technical details
which are not included in the paper due to space limit.

\subsection{Hyperparameters.}
\label{sec:hyper}

\begin{table}[t]
\centering
\begin{tabular}{lc}
\toprule
Hyperparameter & Value\\
\midrule
Optimizer & AdamW\\
Learning Rate & 0.0001 \\
Weight Decay & 0.0001\\
Batch Size & 32 \\
\midrule
Transformer Latent Dimension & $256 $\\
Transformer Heads & $4$\\
Transformer Feedforward Dimension & 1024\\
Transformer Num Layers & 8\\
Dropout & $0.1$\\
\midrule
Diffusion Noise Schedule & Cosine\\
Diffusion Step & $1000$\\
Variance & Fixed Small\\

\bottomrule
\end{tabular}
\caption{
    \textbf{Hyperparameters of PCMDM.}
}
\label{tab:pcmdm}
\end{table}

\begin{table}[t]
\centering
\begin{tabular}{lc}
\toprule
Hyperparameter & Value\\
\midrule
Optimizer & AdamW\\
Learning Rate & 0.0001 \\
Weight Decay & 0.0001\\
Batch Size & 32 \\
\midrule
Motion Transformer Latent Dimension & $512 $\\
Motion Transformer Heads & 8\\
Motion Transformer Feedforward Dimension & $768$\\
Motion Transformer Num Layers & $6$\\
Transformer Heads & $4$\\
Dropout & $0.1$\\
\midrule
Text Encoder & CLIP ViT-B/32\\
\midrule
Temperature of Contrastve Loss& 0.1\\
\bottomrule
\end{tabular}
\caption{
    \textbf{Hyperparameters of MotionCLIP.}
}
\label{tab:motionclip}
\end{table}

To facilitate better replication of our experiment, we have provided detailed hyperparameter settings used in the experiment. Hyperparameters used in PCMDM are listed in Tab~\ref{tab:pcmdm}. Hyperparameters used in MotionClip are listed in Tab~\ref{tab:motionclip}. The former mainly follows the setting of MDM~\cite{tevet2022human}, while the latter mainly follows the setting of Flame~\cite{kim2022flame}.

The architecture of PCMDM has been explained clearly in the main text. MotionCLIP consists of a Motion Encoder (a Transformer Encoder) and a pre-trained CLIP Text Encoder. We encode the action-text pairs in the training set separately with the Motion Encoder and Text Encoder and then use contrastive learning to train MotionCLIP by comparing the encoded results.

\begin{figure}[ht]
\begin{center}
\includegraphics[width=1\linewidth]{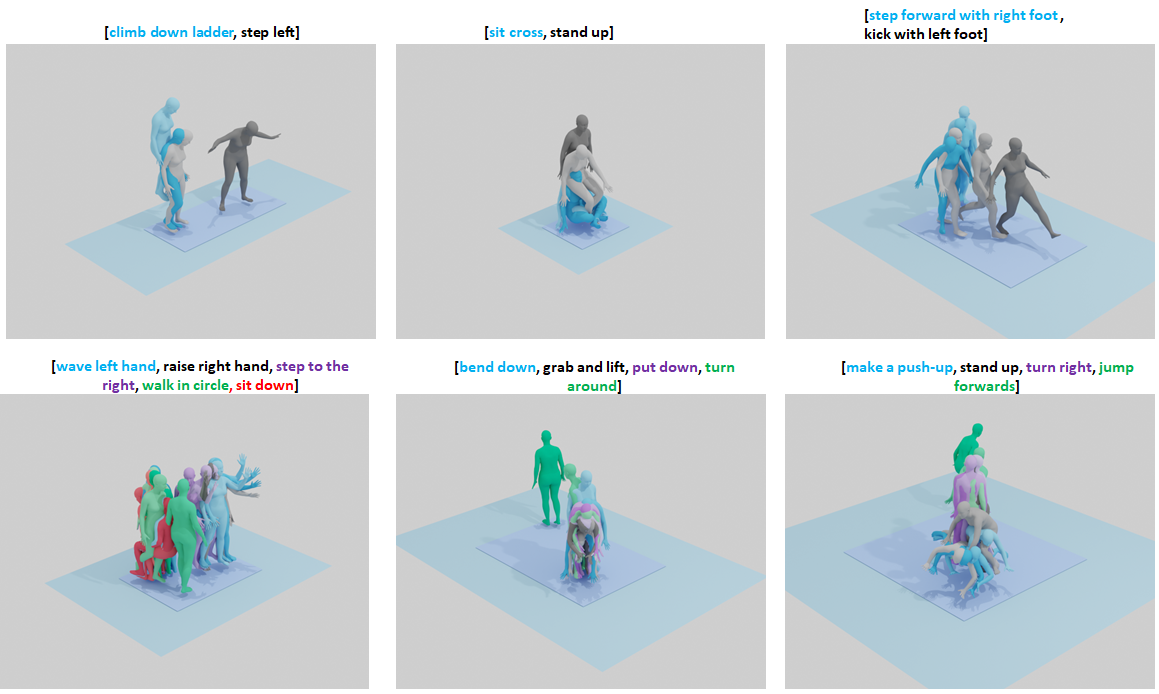}
\caption{\textbf{More visualized results.}}
\label{fig:morevis}
\end{center}
\vskip -0.1in
\end{figure}

\subsection{More visualized results.}
\label{sec:vis}
We provide more visual results in Fig.~\ref{fig:morevis}. Different colored motions represent different text
prompts. We show the overall motion in a single image.

\subsection{Compositional Diffusion Models.}
\label{sec:vis}
In the main text, we mentioned that conditional compositional sampling could be performed using a pre-trained diffusion model~\cite{liu2022compositional}. We will provide further explanation for this. We know that diffusion models are consistent with denoising score matching~\cite{songscore}:
\begin{equation}
\begin{aligned}
    \nabla_x \log q_\sigma(x) \approx-\frac{\epsilon_\theta(x, t)}{\sigma_t}
\end{aligned}
\end{equation}
For the sake of exposition, we will prove this from the perspective of score matching with the score $\nabla_x \log q_\sigma(x)$, and our explanation is mainly based on ~\cite{du2023reduce}. 

For the simple combination distribution of two conditions, i.e. a product model  $q^{\mathrm{prod}}(x) \propto q^1(x) q^2(x)$.  The true score of this distribution is given by:
\begin{equation}
\begin{aligned}
    \nabla_x \log \tilde{q}_t^{\mathrm{prod}}\left(x_t\right)=\nabla_x \log \left(\int d x_0 q^1\left(x_0\right) q^2\left(x_0\right) q\left(x_t \mid x_0\right)\right)
\end{aligned}
\end{equation}
However, this score is difficult to obtain. We can approximate this true score through another simple score summation:
\begin{equation}
\begin{aligned}
    \begin{aligned}
    \nabla_x \log q_t^{\mathrm{prod}}\left(x_t\right) & =\nabla_x \log \left(\int d x_0 q^1\left(x_0\right) q\left(x_t \mid x_0\right)\right) \\
    & +\nabla_x \log \left(\int d x_0 q^2\left(x_0\right) q\left(x_t \mid x_0\right)\right)
    \end{aligned}
\end{aligned}
\end{equation}
This is consistent with the formula for conditional combination derived from the perspective of the diffusion model in~\cite{liu2022compositional}. Note that when $t=0$ we have:
\begin{equation}
\begin{aligned}
    \nabla_x \log \tilde{q}_t^{\mathrm{prod}} = \nabla_x \log q_t^{\mathrm{prod}}\left(x_t\right)
\end{aligned}
\end{equation}
So we can use this sum-based method to approximate the true product distribution.

\begin{table}[ht]
\centering
\caption{Remove Duplicates.}
\label{tab:ood1}
\begin{tabular}{@{}lccc@{}}
\toprule
Method            & FID $\downarrow$ & R-Precision (TOP 3) $\uparrow$ & Diversity $\rightarrow$ \\
\midrule
Real              & $0.018^{\pm0.002}$ & $0.821^{\pm0.002}$ & $15.415^{\pm0.099}$ \\
TEACH             & $8.893^{\pm0.019}$ & $0.931^{\pm0.004}$ & $14.631^{\pm0.051}$ \\
MDM w/ IS         & $7.662^{\pm0.255}$ & $0.824^{\pm0.005}$ & $14.202^{\pm0.054}$ \\
MDM w/ CS         & $7.889^{\pm0.272}$ & $0.830^{\pm0.006}$ & $14.303^{\pm0.045}$ \\
PCMDM w/ IS       & $5.483^{\pm0.111}$ & $0.858^{\pm0.006}$ & $14.827^{\pm0.097}$ \\
PCMDM w/ CS       & $5.215^{\pm0.153}$ & $0.877^{\pm0.004}$ & $14.946^{\pm0.073}$ \\

\bottomrule
\end{tabular}
\end{table}

\begin{table}[ht]
\centering
\caption{Query GPT-3.}
\label{tab:ood2}
\begin{tabular}{@{}lccc@{}}
\toprule
Method            & FID $\downarrow$ & R-Precision (TOP 3) $\uparrow$ & Diversity $\rightarrow$ \\
\midrule
Real              & $0.009^{\pm0.001}$ & $0.297^{\pm0.002}$ & $14.985^{\pm0.042}$ \\
TEACH             & $20.296^{\pm0.483}$ & $0.353^{\pm0.005}$ & $13.501^{\pm0.090}$ \\
MDM w/ IS         & $18.670^{\pm0.386}$ & $0.358^{\pm0.006}$ & $12.775^{\pm0.050}$ \\
MDM w/ CS         & $18.853^{\pm0.209}$ & $0.364^{\pm0.009}$ & $12.849^{\pm0.074}$ \\
PCMDM w/ IS       & $21.062^{\pm0.320}$ & $0.332^{\pm0.008}$ & $13.043^{\pm0.064}$ \\
PCMDM w/ CS       & $20.863^{\pm0.140}$ & $0.333^{\pm0.009}$ & $13.112^{\pm0.057}$ \\
\bottomrule
\end{tabular}
\end{table}
\subsection{Generalization Evaluation.}
While the proposed method demonstrates promising results in generating long-term human motion, the evaluation may be limited to specific datasets or scenarios, which may not fully represent the diversity and complexity of real-world applications. Conducting evaluations on a wider range of datasets and scenarios would strengthen the robustness and generalizability of the proposed method.

Since BABEL is the only dataset suitable for our task, we designed two types of experiments based on BABEL to evaluate the generalization of our method. The first experiment involved exclusively using text streams in the test set that do not overlap with the action phrases present in the training set. The action phrase represents the smallest unit of a single action. For instance, if the action phrase "swing the golf club" appeared in the training set, we would remove all instances of data in the test set that include this particular action phrase. In total, there were 1992 unique phrases in the test set. After eliminating duplicates, the test set contained 1229 unique action phrases that did not appear in the training set. The results of the first experiment are shown in~\ref{tab:ood1}.

It can be observed that the diffusion model methods based on Inpainting Sampling (IS) and Compositional Sampling (CS) did not exhibit a significant decrease in FID, while TEACH [1] experienced a decrease from 7.312 to 8.893. Therefore, it can be concluded that our method demonstrates a significantly better generalization. In the original BABEL [3] dataset, the repeated action phrases are mostly short phrases with fewer words, e.g. "walk" and "stand". After removing duplicate action phrases, the average prompt length in the test set increased. The R-Precision, which is used to evaluate the accuracy of action and text retrieval, improves as the text prompt length increases. Longer text prompts provide more information, leading to an improvement in R-Precision for all methods, including real data.

In the second experiment, we obtained aliases for each action phrase by querying GPT-3 and replacing the text prompts in the test set with these aliases. This task presented a challenge because the original words in the dataset were mostly common words, and GPT-3 had the potential to transform the action phrases into combinations of less common words. For instance, the phrase "rotate left leg to the left behind" could be transformed by GPT-3 into "swivel left leg to the rear." The results of the second experiment are shown in~\ref{tab:ood2}.

Many of the text prompts given by GPT-3 contain uncommon words that our model has not seen in the training dataset. As a result, in such a setting, the performance of all methods sharply declines. The best-performing approach in this scenario is the MDM method based on our proposed Coherent Sampling. MDM [2] does not require a Past Encoder and is trained solely on individual subsequence-text pairs. Because in the case where the text prompt satisfies OOD, the generation of the previous subsequence already shows relatively poor performance. If we continue to encode the poorly generated subsequence using the Past Encoder and feed it to the following subsequence, it will result in error accumulation. Although Coherent Sampling does not show significant improvements in the test metrics, it is still necessary. Because using MDM's standard sampling method can only generate multiple disjointed subsequences, we need to use Coherent Sampling to make the generated sequences sufficiently coherent. The coherent Sampling technique is one of the most important contributions of our work. 

\end{document}